\documentclass[sigconf]{acmart}
\AtBeginDocument{%
  }
\setcopyright{acmlicensed}
\copyrightyear{2025} 
\acmYear{2025} 
\setcopyright{acmlicensed}\acmConference[WWW Companion '25]{Companion
Proceedings of the ACM Web Conference 2025}{April 28-May 2, 2025}{Sydney,
NSW, Australia}
\acmBooktitle{Companion Proceedings of the ACM Web Conference 2025 (WWW
Companion '25), April 28-May 2, 2025, Sydney, NSW, Australia}
\acmDOI{10.1145/3701716.3718372}
\acmISBN{979-8-4007-1331-6/2025/04}
\begin{document}

\title[Knowledge-Decoupled Synergetic Learning]{Knowledge-Decoupled Synergetic Learning: An MLLM based Collaborative Approach to Few-shot Multimodal Dialogue Intention Recognition}

\author{Bin Chen}
\authornote{Both authors contributed equally to this research.}
\affiliation{%
  \institution{University of Chinese Academy of Sciences}
  \city{Hangzhou}
  \state{Zhejiang}
  \country{China}
}
\email{chenbin232@mails.ucas.ac.cn}

\author{Yu Zhang}
\authornotemark[1]
\affiliation{%
  \institution{University of Chinese Academy of Science}
  \city{Hangzhou}
  \state{Zhejiang}
  \country{China}}
\email{zhangyu2312@mails.ucas.ac.cn}

\author{Hongfei Ye}
\affiliation{%
  \institution{University of Chinese Academy of Science}
  \city{Hangzhou}
  \state{Zhejiang}
  \country{China}
  }
\email{yehongfei23@mails.ucas.ac.cn}

\author{Ziyi Huang}
\affiliation{%
 \institution{Zhejiang University}
 \city{Hangzhou}
  \state{Zhejiang}
  \country{China}}
 \email{22360297@zju.edu.cn}

\author{Hongyang Chen}
\authornote{Corresponding author:dr.h.chen@ieee.org}
\affiliation{%
  \institution{Zhejiang Lab}
  \city{Hangzhou}
  \state{Zhejiang}
  \country{China}
}
\email{dr.h.chen@ieee.org}
\renewcommand{\shortauthors}{Bin Chen, Yu Zhang, Hongfei Ye, Ziyi Huang, and Hongyang Chen}
\begin{abstract}
  Few-shot multimodal dialogue intention recognition is a critical challenge in the e-commerce domainn. Previous methods have primarily enhanced model classification capabilities through post-training techniques. However, our analysis reveals that training for few-shot multimodal dialogue intention recognition involves two interconnected tasks, leading to a seesaw effect in multi-task learning. This phenomenon is attributed to knowledge interference stemming from the superposition of weight matrix updates during the training process. To address these challenges, we propose Knowledge-Decoupled Synergetic Learning (KDSL), which mitigates these issues by utilizing smaller models to transform knowledge into interpretable rules, while applying the post-training of larger models. By facilitating collaboration between the large and small multimodal large language models for prediction, our approach demonstrates significant improvements. Notably, we achieve outstanding results on two real Taobao datasets, with enhancements of 6.37\% and 6.28\% in online weighted F1 scores compared to the  state-of-the-art method, thereby validating the efficacy of our framework.
\end{abstract}

\maketitle

\section{Introduction}
In the e-commerce sector, various investigations have explored specialized approaches for comprehending customer intentions\cite{zhao2019dynamic,qiu2022pre,ahmadvand2020jointmap}, ranging from deep learning models for purchase intention recognition \cite{ma2024identifying} to multimodal dialogue systems that synthesize both textual and visual information \cite{yuan2022mcic}. While these contributions represent significant advancements, they predominantly concentrate on general dialogue scenarios or single-modality approaches. The complexity of managing diverse modalities, such as text and images, alongside the need to balance performance across multiple tasks, presents substantial challenges for few-shot multimodal dialogue intention recognition within e-commerce customer service systems.

\begin{figure}[h]\scriptsize
	\begin{center}
		\tabcolsep 1pt
			\includegraphics[width = 0.85\columnwidth]{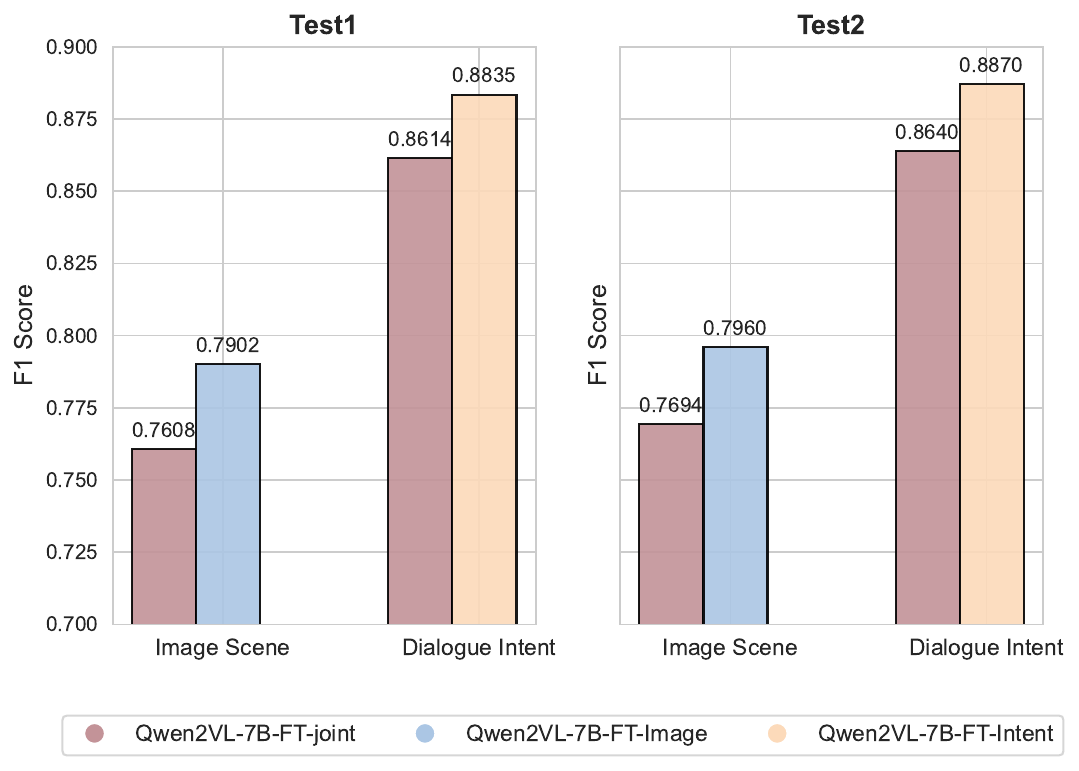}              \\
	\end{center}	
	\caption{The \textbf{Qwen2VL$_{7B}$-FT-Image} and \textbf{Qwen2VL$_{7B}$-FT-Intent} models specialize in image and intent classification, respectively, using full fine-tuning. In contrast, \textbf{Qwen2VL$_{7B}$-FT-joint} is jointly trained for both tasks. Performance comparisons on Test Set 1 and Test Set 2 from Taobao show that single-task models outperform joint models with identical parameters, suggesting a seesaw effect in multimodal intent recognition for e-commerce applications.}	
	\label{fig:phi}
\end{figure}

With the rapid advancement of multimodal large language models (MLLMs)\cite{driess2023palm,zhang2024mm,liang2024survey,xia2025llmga}, such as LLaVA\cite{liu2024visual}, QwenVL\cite{bai2023qwen1,bai2023qwen2,wang2024qwen2}, and InternVL\cite{chen2024internvl}, these sophisticated frameworks have been widely applied to downstream tasks. However, general-purpose multimodal large language models often lack the specific knowledge pertinent to e-commerce scenarios, typically relying on post-training methods to enhance their performance\cite{liu2021post,shang2023post,xiao2023smoothquant}. Our investigation reveals that training for multimodal dialogue intention recognition involves two interconnected tasks, leading to a seesaw effect in multi-task learning—where performance improvements in one task can result in declines in others, as illustrated in Figure \ref{fig:phi}. Through experimental analysis, we identify that this phenomenon is primarily due to knowledge interference caused by the superposition of weight matrix updates during training. This interference poses significant limitations when handling multimodal and multi-task settings, as the model struggles to optimize multiple objectives simultaneously.
To tackle these challenges, we propose Knowledge-Decoupled Synergetic Learning (KDSL), a novel framework aimed at mitigating knowledge interference and task imbalance in multimodal and multi-task learning scenarios. The central premise of KDSL is to decouple the common-sense knowledge representations acquired through extensive corpus training and contextually learned representations from the non-interpretable parameter space of MLLMs, transferring them into a more actionable semantic space. Our approach involves utilizing a smaller MLLM to employ Monte Carlo tree search strategies for generating and collecting rules, which are stored in a rule base, thereby achieving a separation of knowledge and model parameters. Compared to traditional Monte Carlo tree search methods\cite{chaslot2010monte}, we streamline the simulation and evaluation processes by constructing multiple agents from the same model for simulation and enabling agents to self-assess, thereby enhancing search efficiency. Additionally, KDSL involves fine-tuning the MLLM using the Taobao few-shot multimodal dialogue intention dataset with data augmentation. This approach enhances the model's capacity to learn implicit patterns that are often difficult to capture through explicit rules. Finally, predictions are made through a collaborative approach between the large and small Multimodal language models. KDSL effectively resolves the knowledge interference issues associated with multi-task post-training. Our method achieves remarkable results on two real Taobao datasets, with improvements of 6.37\% and 6.28\% in online weighted F1 scores compared to the state-of-the-art method, thereby demonstrating the efficacy of our approach.
In summary, this work makes three primary contributions:
\begin{itemize}
    \item We identify the seesaw effect in Multimodal Large Language Models (MLLMs) when learning few-shot multimodal intention recognition tasks in e-commerce, and we analyze the underlying cause of this phenomenon—knowledge interference due to training weight updates.
    
    \item We propose a method that leverages Monte Carlo tree search to enhance the rule generation capabilities of smaller MLLMs, simplifying the search process by enabling self-assessment within the model.
    
    \item We introduce KDSL, a novel framework that integrates a rule library generated by smaller MLLMs with post-training of larger MLLMs for collaborative prediction. Our method achieves outstanding results in multimodal dialogue intention recognition, yielding improvements of 6.37\% and 6.28\% in online weighted F1 scores compared to the state-of-the-art methods, thereby demonstrating its efficacy.
\end{itemize}
\section{Method}
\begin{figure*}[!t]\scriptsize
	\begin{center}
		\tabcolsep 1pt
			\includegraphics[width = 0.7\textwidth]{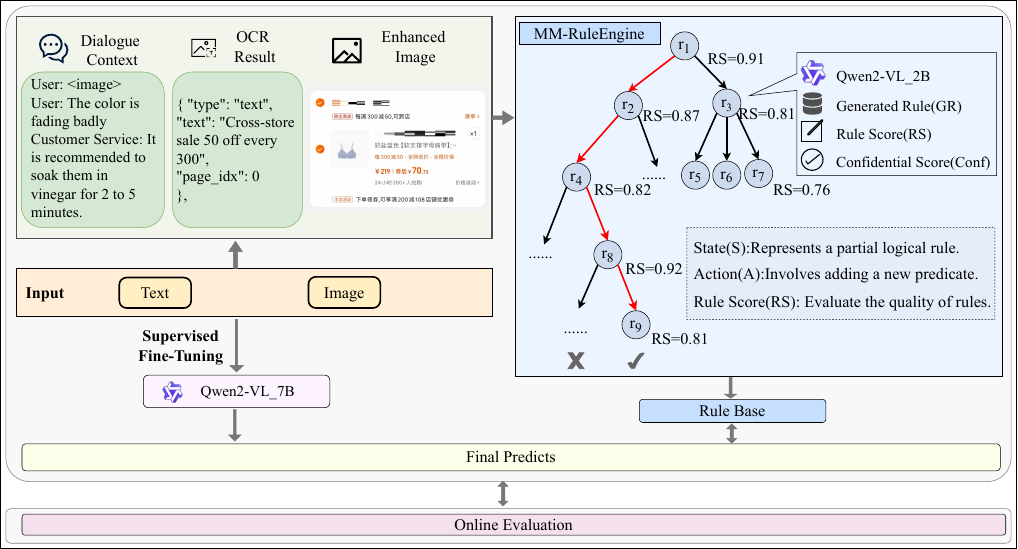}              \\
		
	\caption{\textbf{
    We propose a collaborative pipeline integrating multimodal large and small language models. The large model, Qwen2-VL$_{7B}$, is fine-tuned on the Taobao few-shot multimodal dialogue intention dataset with data augmentation, enabling it to learn implicit patterns. The smaller model, Qwen2-VL$_{2B}$, uses Monte Carlo Tree Search to generate and collect rules, which are stored in a rule base. The fine-tuned Qwen2-VL$_{7B}$ and the rule base collaborate for prediction.
    } }	
	\label{fig:overall}\end{center}
\end{figure*}
As shown in Figure 2, we propose a collaborative pipeline combining multimodal large and small language models. We fine-tune the large models using the Taobao few-shot multimodal dialogue intention dataset with data augmentation, allowing the model to learn implicit patterns. The small models employ Monte Carlo tree search for rule generation, which are stored in a rule base, decoupling knowledge from model parameters. Finally, the fine-tuned large model and rule base collaborate for prediction.

\subsection{MM Rule-Engine Construction}
We construct the rule engine is constructed through a multi-stage pipeline, which includes data processing, rule generation, filtering and evaluation. 
\textbf{Data Processing:} We employ LayoutLMv3\cite{huang2022layoutlmv3} to extract OCR information and separate user and customer utterances in multi-turn dialogues. OCR data supports rule generation for image classification, while dialogue utterances aid intent classification. A comparison on 1,000 samples shows LayoutLMv3 outperforms Qwen2VL$_{2B}$, with only 15.4\% of its OCR outputs being invalid or empty. However, its limited OCR capability necessitates combining rules with fine-tuned domain models for comprehensive performance. For rule generation, training samples with the same label serve as input, while validation data is generated via rephrased training examples to ensure consistency.

\textbf{Rule Generation:} We employ a Monte Carlo Tree Search (MCTS)-based framework integrated with MLLMs to generate rules, addressing challenges such as reliance on external rewards and the high cost of repeated simulations. In this framework, the MLLM is virtualized as an agent that, based on the current state $S$, proposes actions $A$ (adding predicates to rules), evaluates the resulting rules with a reward score $R$ and provides a confidence score for the evaluation. These reward scores reflect the accuracy of the generated rules on validation examples presented in the MLLM-generated context. To expand the search tree, sub-agents calculate Upper Confidence Bound applied to Trees (UCT) values, selecting the node with the highest UCT score for further exploration. The simulation terminates when the number of predicates in a rule exceeds a predefined threshold (5), ensuring computational efficiency. Each state $S$ corresponds to a partial logical rule, initialized as $S_0 = \emptyset$ and expanded iteratively by adding predicates $\alpha_i$, resulting in $S_n = S_{n-1} \cup \{\alpha_i\}$. The action space $A$ consists of candidate predicates $\alpha_i$ generated by the MLLM, which are added to the current state. The reward function $R$ evaluates the quality of the logical rules using validation examples provided in the MLLM’s context. The reward score, along with its confidence, serves as the initial reward for subsequent computations, guiding the rule refinement process.

\textbf{Rule Filtering and Evaluation:} After rule generation, we filter out rules with reward scores below 0.8 to ensure quality, balancing the number of rules with their overall quality. For overlapping rules, if one rule’s predicates form a subset of another’s and the former has a higher reward score, the latter is removed. We then apply online validation to eliminate erroneous rules, ensuring robustness. 
\subsection{Enhancing E-commerce Knowledge Understanding in Multimodal Large Language Models}
Rule-based systems demonstrate limited capability in capturing latent semantic relationships for multimodal dialogue intention recognition. Fine-tuning large models is essential for acquiring domain-specific knowledge, such as distinguishing logistics detail pages from tracking pages in e-commerce scenarios. Accordingly, we fine-tuned Qwen2VL$_{7B}$ on a Taobao few-shot multimodal dialogue intention dataset.

Additionally, we employed data augmentation techniques to
improve the model’s generalization capability. Our experiments
included various combinations of data augmentation, such as multiturn dialogue text augmentation, pixel and spatial structure transformations on images, and their combinations. To balance model
accuracy and training time, we ultimately adopted horizontal flipping as the primary data augmentation strategy during training.
\subsection{MM Rule-Engine and Multimodal Large Language Model for Collaborative Reasoning}
We observe that different model categories demand varying numbers of training steps for convergence and that multi-task learning often exhibits a see-saw effect. Meanwhile, the visual encoder shows limited capacity to discriminate fine-grained image differences. Rule-based engines excel in explicitly handling edge cases, particularly when the main model is prone to errors, but their coverage remains constrained. To leverage the strengths of both, we combine an MM Rule-Engine with our fully fine-tuned Qwen2VL$_{7B}$ model. We feed inputs to both components and use MM Rule-Engine outputs to correct Qwen2VL$_{7B}$ predictions, thereby creating a complementary synergy that enhances overall performance.
\section{Experiments}
We conduct comparative and ablation experiments to validate our pipeline. We first describe the setup, then compare results with strong baselines, and finally analyze each module’s contribution.
\subsection{Experimental Setup and Baselines}
\label{sec:ExperimentalSetup}

Experiments are conducted on 2 NVIDIA A100 GPUs with Ubuntu 20.04. For Qwen2VL fine-tuning, we set a cutoff length of 1024, a learning rate of 5.0e-6, and train for 3 epochs to balance performance and efficiency.
We evaluate several baseline models. \textbf{Qwen2VL$_{7B}$-FT} is fully fine-tuned for joint multimodal dialogue intention recognition and image recognition, while \textbf{Qwen2VL$_{7B}$-COT-FT} incorporates chain-of-thought (CoT) data construction. \textbf{Qwen2VL$_{2B}$-FT} is a smaller 2B-parameter variant, and \textbf{InternVL$_{8B}$-FT} employs an 8B-parameter InternVL architecture for multimodal learning. Our methods, \textbf{Qwen2VL$_{7B}$-FT$_{aug}$} and \textbf{KDSL}, integrate data augmentation and rule-based reasoning to enhance performance.
\subsection{Datasets and Evaluation Metrics}
\label{sec:DatasetsAndMetrics}

We use two E-commerce datasets from Taobao e-commerce scenarios. Test Set 1 consists of 10,000 samples evenly split between multimodal dialogue intent recognition and image scene recognition. Test Set 2 also has 10,000 samples, with 3,674 for multimodal dialogue intent recognition and 6,326 for image scene recognition, ensuring diverse evaluation.

Performance is measured using class-weighted F1 scores. \textbf{Dialogue Intention Score} (DIS) evaluates multimodal dialogue intent recognition, \textbf{Image Scene Score} (ISS) assesses image recognition, and \textbf{Online Submit Score} offers a comprehensive multimodal evaluation by unifying the assessment of both multimodal dialogue intent recognition and image recognition.

\section{Results}
\begin{table}[htbp]
\centering
\renewcommand{\arraystretch}{1} 
{
\resizebox{\columnwidth}{!}{%
\begin{tabular}{@{}lccc|ccc@{}}
\toprule
\textbf{Methods} & \multicolumn{3}{c|}{\textbf{Test Set 1}} & \multicolumn{3}{c}{\textbf{Test Set 2}} \\
\cmidrule(lr){2-4} \cmidrule(lr){5-7}
 & \textbf{DIS} & \textbf{ISS} & \textbf{OSS} & \textbf{DIS} & \textbf{ISS} & \textbf{OSS} \\
\midrule
Qwen2VL$_{7B}$-FT           & 0.8614 & 0.7608 & 0.8111 & 0.8640 & 0.7694 & 0.8360 \\
Qwen2VL$_{7B}$-COT-FT      & 0.8462 & 0.7431 & 0.7947 & 0.8503 & 0.7409 & 0.8179 \\
Qwen2VL$_{2B}$-FT           & 0.8620 & 0.7128 & 0.7874 & 0.8664 & 0.7192 & 0.8229 \\
InternVL$_{8B}$-FT          & 0.8858 & 0.7703 & 0.8281 & 0.8898 & 0.7688 & 0.8540 \\
Qwen2VL$_{7B}$-FT$_{aug}$(ours) & \underline{0.8920} & \underline{0.7767} & \underline{0.8343} & \underline{0.8960} & \underline{0.7900} & \underline{0.8646} \\
KDSL (ours)                    & \textbf{0.9122} & \textbf{0.8373} & \textbf{0.8748} & \textbf{0.9273} & \textbf{0.8340} & \textbf{0.8988} \\
\bottomrule
\end{tabular}%
}}
\caption{Performance scores for various methods across different tasks and test sets. Test Set 1 and Test Set 2 results are reported across three metrics: DIS, ISS, and OSS. The best results are in bold, and the second best are \underline{underlined}.}
\label{tab:f1_scores}
\vspace{-20pt}
\end{table}
Table 1 compares the performance of different methods across three key evaluation metrics: \textbf{Dialogue Intention Score}(DIS), \textbf{Image Scene Score}(ISS), and \textbf{Online Submit Score}(OSS), evaluated on two test sets. The results demonstrate the effectiveness of our proposed methods, \textbf{Qwen2VL$_{7B}$-FT$_{aug}$} and \textbf{KDSL}, in achieving state-of-the-art performance. The baseline method, \textbf{Qwen2VL$_{7B}$-FT}, achieves DIS of 86.14\% and 86.40\%, Image Scene Scores of 76.08\% and 76.94\%, and Online Submit Scores of 81.11\% and 83.60\%. \textbf{InternVL$_{8B}$-FT} surpasses it with improvements of 2.44\% and 2.58\% in DIS and 1.70\% and 1.80\% in Online Submit Scores, indicating greater benefits for dialogue understanding. Conversely, \textbf{Qwen2VL$_{2B}$-FT} underperforms, with declines of 4.8\% and 5.02\% in ISS, emphasizing the impact of model scale. \textbf{Qwen2VL$_{7B}$-COT-FT}, with chain-of-thought (CoT) fine-tuning, shows reduced performance, with decreases of 1.52\% and 1.37\% in DIS and 1.77\% and 2.85\% in ISS, suggesting limited effectiveness in multimodal tasks. In contrast, \textbf{Qwen2VL$_{7B}$-FT$_{aug}$}, leveraging data augmentation, improves DIS by 3.06\% and 3.20\% and ISS by 1.59\% and 2.06\%, demonstrating its effectiveness in two tasks.
Finally, the \textbf{KDSL} framework, which integrates fine-tuned MLLM using data augmentation with the MM Rule-Engine, achieves superior performance. Specifically, it demonstrates gains of 5.08\% and 6.33\% in DIS, 7.65\% and 6.46\% in ISS, and 6.37\% and 6.28\% in Online Submit Scores. These results underscore the complementary strengths of fine-tuned MLLM with data augmentation and structured reasoning.
\section{Ablation Study and Analysis}
\label{sec:ablation}
To analyze the impact of each component, we conduct ablation experiments on (1) MCTS-based rule generation and (2) data augmentation. We evaluate two variants:

\begin{enumerate}
  \item \textbf{FT-MLLM}: FT-MLLM is fine-tuned solely on the training dataset, excluding both data augmentation and the MM Rule-Engine.
  \item \textbf{FT-MLLM-dataaug}: Removes MM Rule-Engine to assess the effect of data augmentation.
  \item \textbf{FT-MLLM-RE}: Omits data augmentation to isolate the impact of MM Rule-Engine.
\end{enumerate}
\begin{table}[htbp]
\centering
\resizebox{\columnwidth}{!}{%
\begin{tabular}{@{}lccc|ccc@{}}
\toprule
\textbf{Methods} & \multicolumn{3}{c|}{\textbf{Test Set 1}} & \multicolumn{3}{c}{\textbf{Test Set 2}} \\
\cmidrule(lr){2-4} \cmidrule(lr){5-7}
 & \textbf{DIS} & \textbf{ISS} & \textbf{OSS} & \textbf{DIS} & \textbf{ISS} & \textbf{OSS} \\
\midrule
FT-MLLM            & 0.8614 & 0.7608 & 0.8111 & 0.8640 & 0.7694 & 0.8360 \\
FT-MLLM-dataaug    & \underline{0.8920} & 0.7767 & 0.8343 & 0.8960 & 0.7900 & 0.8646 \\
FT-MLLM-RE         & 0.8820 & \underline{0.8219} & \underline{0.8520} & \underline{0.9138} & \underline{0.8220} & \underline{0.8866} \\
KDSL             & \textbf{0.9122} & \textbf{0.8373} & \textbf{0.8748} & \textbf{0.9273} & \textbf{0.8340} & \textbf{0.8988} \\
\bottomrule
\end{tabular}%
}
\caption{Ablation study results illustrating the impact of different components on performance metrics. Test Set 1 and Test Set 2 results are reported across three metrics: DIS, ISS, and OSS. The best results are in bold, and the second best are \underline{underlined}.}
\label{tab:ablation_study}
\vspace{-20pt}
\end{table}

Table \ref{tab:ablation_study} presents a comparative analysis of model performance across Test Set 1 and Test Set 2. The baseline method, FT-MLLM, achieves scores of 76.08\% and 76.94\%, while data augmentation (FT-MLLM-dataaug) improves performance to 77.67\% and 79.00\%, highlighting its role in enhancing robustness.

FT-MLLM-RE, incorporating the MM Rule-Engine, achieves substantial gains of 4.09\% (85.20\%) on Test Set 1 and 5.06\% (88.66\%) on Test Set 2, demonstrating the benefits of structured reasoning. The proposed KDSL method, combining data augmentation and MM Rule-Engine, further improves scores to 87.48\% and 89.88\%, achieving relative increases of 6.37\% and 6.28\% over FT-MLLM. These results confirm the complementary effects of fine-tuned MLLM with data augmentation and MM Rule-Engine in optimizing few-shot multimodal dialogue intention recognition performance.
\section{Conclusion}
This paper identifies the seesaw phenomenon in few-shot multimodal dialogue intention recognition tasks within e-commerce and introduces Knowledge-Decoupled Synergetic Learning (KDSL) to address this challenge. Experimental results on two real Taobao datasets demonstrate that KDSL achieves improvements of 6.37\% and 6.28\% in online weighted F1 scores, validating the effectiveness of our framework.
\begin{acks}
This work is supported in part by National Key Research and Development Program of China (2022YFB4500300), in part by National Natural Science Foundation of China under Grant 62271452.
\end{acks}





\appendix
\bibliographystyle{acm}
\balance
\bibliography{software}

\begin{thebibliography}{10}

\bibitem{ahmadvand2020jointmap}
{\sc Ahmadvand, A., Kallumadi, S., Javed, F., and Agichtein, E.}
\newblock Jointmap: joint query intent understanding for modeling intent hierarchies in e-commerce search.
\newblock In {\em Proceedings of the 43rd International ACM SIGIR Conference on Research and Development in Information Retrieval\/} (2020), pp.~1509--1512.

\bibitem{bai2023qwen1}
{\sc Bai, J., Bai, S., Yang, S., Wang, S., Tan, S., Wang, P., Lin, J., Zhou, C., and Zhou, J.}
\newblock Qwen-vl: A frontier large vision-language model with versatile abilities.
\newblock {\em arXiv preprint arXiv:2308.12966\/} (2023).

\bibitem{bai2023qwen2}
{\sc Bai, J., Bai, S., Yang, S., Wang, S., Tan, S., Wang, P., Lin, J., Zhou, C., and Zhou, J.}
\newblock Qwen-vl: A versatile vision-language model for understanding, localization, text reading, and beyond.
\newblock {\em arXiv preprint arXiv:2308.12966 1}, 2 (2023), 3.

\bibitem{chaslot2010monte}
{\sc Chaslot, G. M. J.-B.~C.}
\newblock Monte-carlo tree search.

\bibitem{chen2024internvl}
{\sc Chen, Z., Wu, J., Wang, W., Su, W., Chen, G., Xing, S., Zhong, M., Zhang, Q., Zhu, X., Lu, L., et~al.}
\newblock Internvl: Scaling up vision foundation models and aligning for generic visual-linguistic tasks.
\newblock In {\em Proceedings of the IEEE/CVF Conference on Computer Vision and Pattern Recognition\/} (2024), pp.~24185--24198.

\bibitem{driess2023palm}
{\sc Driess, D., Xia, F., Sajjadi, M.~S., Lynch, C., Chowdhery, A., Ichter, B., Wahid, A., Tompson, J., Vuong, Q., Yu, T., et~al.}
\newblock Palm-e: an embodied multimodal language model.
\newblock In {\em Proceedings of the 40th International Conference on Machine Learning\/} (2023), pp.~8469--8488.

\bibitem{huang2022layoutlmv3}
{\sc Huang, Y., Lv, T., Cui, L., Lu, Y., and Wei, F.}
\newblock Layoutlmv3: Pre-training for document ai with unified text and image masking.
\newblock In {\em Proceedings of the 30th ACM International Conference on Multimedia\/} (2022), pp.~4083--4091.

\bibitem{liang2024survey}
{\sc Liang, Z., Xu, Y., Hong, Y., Shang, P., Wang, Q., Fu, Q., and Liu, K.}
\newblock A survey of multimodel large language models.
\newblock In {\em Proceedings of the 3rd International Conference on Computer, Artificial Intelligence and Control Engineering\/} (2024), pp.~405--409.

\bibitem{liu2024visual}
{\sc Liu, H., Li, C., Wu, Q., and Lee, Y.~J.}
\newblock Visual instruction tuning.
\newblock {\em Advances in neural information processing systems 36\/} (2024).

\bibitem{liu2021post}
{\sc Liu, Z., Wang, Y., Han, K., Zhang, W., Ma, S., and Gao, W.}
\newblock Post-training quantization for vision transformer.
\newblock {\em Advances in Neural Information Processing Systems 34\/} (2021), 28092--28103.

\bibitem{ma2024identifying}
{\sc Ma, J., Guo, X., and Zhao, X.}
\newblock Identifying purchase intention through deep learning: analyzing the q \&d text of an e-commerce platform.
\newblock {\em Annals of Operations Research 339}, 1 (2024), 329--348.

\bibitem{qiu2022pre}
{\sc Qiu, Y., Zhao, C., Zhang, H., Zhuo, J., Li, T., Zhang, X., Wang, S., Xu, S., Long, B., and Yang, W.-Y.}
\newblock Pre-training tasks for user intent detection and embedding retrieval in e-commerce search.
\newblock In {\em Proceedings of the 31st ACM International Conference on Information \& Knowledge Management\/} (2022), pp.~4424--4428.

\bibitem{shang2023post}
{\sc Shang, Y., Yuan, Z., Xie, B., Wu, B., and Yan, Y.}
\newblock Post-training quantization on diffusion models.
\newblock In {\em Proceedings of the IEEE/CVF conference on computer vision and pattern recognition\/} (2023), pp.~1972--1981.

\bibitem{wang2024qwen2}
{\sc Wang, P., Bai, S., Tan, S., Wang, S., Fan, Z., Bai, J., Chen, K., Liu, X., Wang, J., Ge, W., et~al.}
\newblock Qwen2-vl: Enhancing vision-language model's perception of the world at any resolution.
\newblock {\em arXiv preprint arXiv:2409.12191\/} (2024).

\bibitem{xia2025llmga}
{\sc Xia, B., Wang, S., Tao, Y., Wang, Y., and Jia, J.}
\newblock Llmga: Multimodal large language model based generation assistant.
\newblock In {\em European Conference on Computer Vision\/} (2025), Springer, pp.~389--406.

\bibitem{xiao2023smoothquant}
{\sc Xiao, G., Lin, J., Seznec, M., Wu, H., Demouth, J., and Han, S.}
\newblock Smoothquant: Accurate and efficient post-training quantization for large language models.
\newblock In {\em International Conference on Machine Learning\/} (2023), PMLR, pp.~38087--38099.

\bibitem{yuan2022mcic}
{\sc Yuan, S., Shen, X., Zhao, Y., Liu, H., Yan, Z., Liu, R., and Chen, M.}
\newblock Mcic: multimodal conversational intent classification for e-commerce customer service.
\newblock In {\em CCF International Conference on Natural Language Processing and Chinese Computing\/} (2022), Springer, pp.~749--761.

\bibitem{zhang2024mm}
{\sc Zhang, D., Yu, Y., Dong, J., Li, C., Su, D., Chu, C., and Yu, D.}
\newblock Mm-llms: Recent advances in multimodal large language models.
\newblock {\em arXiv preprint arXiv:2401.13601\/} (2024).

\bibitem{zhao2019dynamic}
{\sc Zhao, J., Chen, H., and Yin, D.}
\newblock A dynamic product-aware learning model for e-commerce query intent understanding.
\newblock In {\em Proceedings of the 28th ACM International Conference on Information and Knowledge Management\/} (2019), pp.~1843--1852.

\end{thebibliography}
\end{document}